\documentclass[conference]{IEEEtran}
\selectfont
\IEEEoverridecommandlockouts
\usepackage[table,xcdraw]{xcolor}
\usepackage{amsmath, enumitem}
\usepackage{amssymb}
\usepackage{amsfonts}
\usepackage{cite}
\usepackage{framed}
\usepackage{graphicx}
\usepackage{textcomp}
\usepackage{xcolor}
\usepackage{multirow}
\usepackage{soul}
\usepackage{tikz}
\usepackage{algorithm}
\usepackage{algpseudocode}
\usepackage{colortbl}
\usepackage{bm}
\usepackage[hyphens]{url}
\usepackage{float}
\usepackage{hyperref}
\usepackage{mdframed}

\usepackage{booktabs}
\usepackage{tabularx}
\usepackage{array}
\newcommand{\cmmnt}[1]{}

\usepackage{pifont}
\newcommand{\cmark}{\ding{51}}%
\newcommand{\xmark}{\ding{55}}%

\usepackage[a4paper, total={184mm,239mm}]{geometry}
\def\BibTeX{{\rm B\kern-.05em{\sc i\kern-.025em b}\kern-.08em
    T\kern-.1667em\lower.7ex\hbox{E}\kern-.125emX}}

\newcommand{\help}[1]{\textcolor{red}{#1}}
    
\begin{document}

\title{EEG2MCU: Sensing-to-Execution Co-Design for Wearable EEG Inference\vspace{-4mm}}

\author{
\IEEEauthorblockN{
        Shoma Yukawa\IEEEauthorrefmark{1},
        Meghna Roychowdhury\IEEEauthorrefmark{2},
        Arnab Sanyal\IEEEauthorrefmark{3},
        Chengwei Zhou\IEEEauthorrefmark{1},
        Shreyas Sen\IEEEauthorrefmark{2}, and
        Gourav Datta\IEEEauthorrefmark{1}
    }
    \vspace{2mm} 
    \IEEEauthorblockA{
        \begin{minipage}[t]{0.33\textwidth}
            \centering
            \IEEEauthorrefmark{1}Case Western Reserve University,\\Cleveland, OH\\
            \texttt{\footnotesize{\{\href{mailto:shoma.yukawa@case.edu}{shoma.yukawa}, \href{mailto:chengwei.zhou@case.edu}{chengwei.zhou}, \href{mailto:gourav.datta@case.edu}{gourav.datta}\}@case.edu}}
        \end{minipage}
        \hfill 
        \begin{minipage}[t]{0.33\textwidth}
            \centering
            \IEEEauthorrefmark{2}Purdue University,\\West Lafayette, IN\\
            \texttt{\footnotesize{\{\href{mailto:mroycho@purdue.edu}{mroycho}, \href{mailto:shreyas@purdue.edu}{shreyas}\}@purdue.edu}}
        \end{minipage}
        \hfill
        \begin{minipage}[t]{0.33\textwidth}
            \centering
            \IEEEauthorrefmark{3}The University of Texas,\\Austin, TX\\
            \texttt{\footnotesize{\href{mailto:sanyal@utexas.edu}{sanyal@utexas.edu}}}
        \end{minipage}
    }
    \vspace{-9mm}
} 

\maketitle

\begin{abstract}
Wearable EEG inference is constrained not only by model size, but also by the channel-dependent sensing front end that must process high-dimensional inputs before compact downstream computation. We present \textbf{EEG2MCU}, a sensing-to-execution co-design framework that treats electrode count as a hardware design variable through task-aware montage selection. EEG2MCU retains a compact time--frequency representation and generates only the INT8 regions required by each first-layer tile, avoiding full input materialization. Halo reuse and access-aware placement across Flash, SRAM, and tightly coupled memory further reduce redundant generation and data movement. On a Cortex-M7 MCU, EEG2MCU reduces peak SRAM by 11.38$\times$ relative to the baseline, including a 1.71$\times$ reduction from the 62-to-19 electrode montage, while reducing latency by 2.57$\times$ and energy by 2.48$\times$. It retains 94--96\% of full-montage accuracy, demonstrating practical high-dimensional EEG inference under tight embedded-resource constraints. 
\end{abstract}


\vspace{-4mm}
\section{Introduction}
\label{sec:introduction}
\vspace{-1mm}

Wearable electroencephalography (EEG) supports continuous neurological
monitoring, seizure detection, affective computing, and brain--computer
interfaces (BCIs)~\cite{he2023wearableeeg}. In wearable neurotechnology and
Internet-of-Medical-Things (IoMT) systems, running inference on the sensing
device itself reduces wireless communication, response latency, privacy
exposure, and battery drain. Yet EEG models have grown from compact CNNs such
as EEGNet and DeepConvNet~\cite{lawhern2018eegnet,schirrmeister2017deep} to
convolution--attention, self-supervised, foundation, and state-space
architectures~\cite{song2023eegconformer,kostas2021bendr,jiang2024labram,
wang2024eegpt,wei2025neurolm,tegon2026femba}, widening the mismatch with the
SRAM, bandwidth, compute, and energy budgets of wearable-class
microcontrollers (MCUs).

Prior work narrows this gap through compact architectures, knowledge
distillation, channel reduction, and optimized MCU
kernels~\cite{ingolfsson2020eegtcnet,baghersalimi2024m2skd,
lin2020mcunet,lai2018cmsisnn}, but optimizes the network only after the
sensing interface and model input are fixed. This hides the dominant cost for
EEG: a high-density recording spans many electrodes, and the resulting
multi-channel time--frequency representation must be generated, stored,
moved, and consumed by the front end. A model with a modest parameter count
thus becomes undeployable once its first layer encounters a
channel-dependent working set that exceeds the memory budget or dominates
execution time. We argue that EEG deployment must be optimized from the
\emph{sensing interface} onward, treating electrode dimensionality as a
design variable rather than a fixed property of the input.

Our layer-wise analysis reveals a strongly \emph{front-loaded} profile: the
channel-dependent first layer holds the largest activation working set and
consumes a disproportionate share of computation and latency, while later
layers operate on progressively compressed representations. This is not
architecture-specific. CNN, Transformer, and state-space EEG models alike
open with a convolutional stem that reads every electrode at full resolution,
so all meet the same working set before their backbones diverge. The front
end is therefore a \emph{feasibility gateway}: deployment requires that the
peak resident working set fit in SRAM and that end-to-end execution,
including representation generation, finish within the EEG window. 
This makes deployment a joint memory--latency
feasibility problem that backbone compression alone cannot solve.

\begin{figure}
    \centering
    \includegraphics[width=1\linewidth]{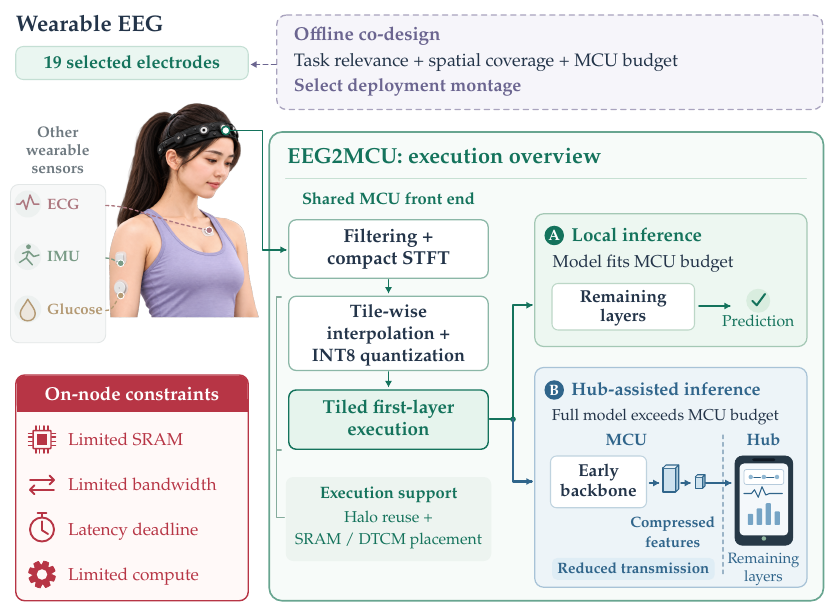}
    \vspace{-8mm}
    \caption{\footnotesize \textbf{EEG2MCU system overview}. Offline co-design selects a hardware-feasible 19-electrode montage, while the wearable MCU streams the compact STFT through tiled first-layer execution. 
    }
    \label{fig:placeholder-i}
    \vspace{-5mm}
\end{figure}

To mitigate this, we propose \textbf{EEG2MCU}, a sensing-to-execution co-design
framework (see Fig.~\ref{fig:placeholder-i}). Our electrode-level sensitivity analysis derives a
deployment-oriented montage under spatial-coverage and hardware constraints,
reducing the channels acquired and processed; unlike post-acquisition feature
pruning, this acts at the sensing interface and so lowers input
dimensionality, first-layer computation, and memory traffic at once. We
further avoid materializing the expanded time--frequency tensor: the compact short-time Fourier transform (STFT) representation stays resident while we fuse interpolation and INT8
quantization with first-layer execution, generating only the rows the current
tile requires. Memory-constrained tiling makes the layer executable under
tight SRAM limits, and halo reuse with access-aware placement across Flash,
SRAM, and tightly coupled memory cuts redundant generation and data movement.
Because the stem is shared, these gains transfer across backbones; models
still exceeding the node budget can offload remaining layers to a nearby
hub~\cite{distnn}.
Our main contributions are as follows:
\begin{itemize}[leftmargin=*]
    \item \textbf{First-layer MCU characterization.}
    We identify the electrode-dependent first layer as the dominant memory
    and latency bottleneck across all evaluated EEG model families, reaching
    87.1\% of end-to-end latency and 80.5\% of computation at 62 electrodes.

    \item \textbf{Saliency-guided montage co-design.}
    We treat electrode configuration as an algorithm--hardware design
    variable; our sensitivity analysis derives a deployment montage under
    spatial-coverage and MCU constraints, cutting front-end computation
    $3.26\times$ at 19 of 62 electrodes while retaining 94--96\% of
    full-montage accuracy.

    \item \textbf{Streaming \& locality-aware execution.}
    We jointly optimize on-demand generation, INT8 tiling, halo and buffer
    reuse, and hierarchy-aware scheduling, cutting peak SRAM by 66\%
    (1374.4\,KiB to 464.9\,KiB) and reducing latency by 10.5\% and energy by
    13.0\% on a Cortex-M7 MCU.

\end{itemize}


\begin{figure}
    \centering
    \includegraphics[width=1\linewidth]{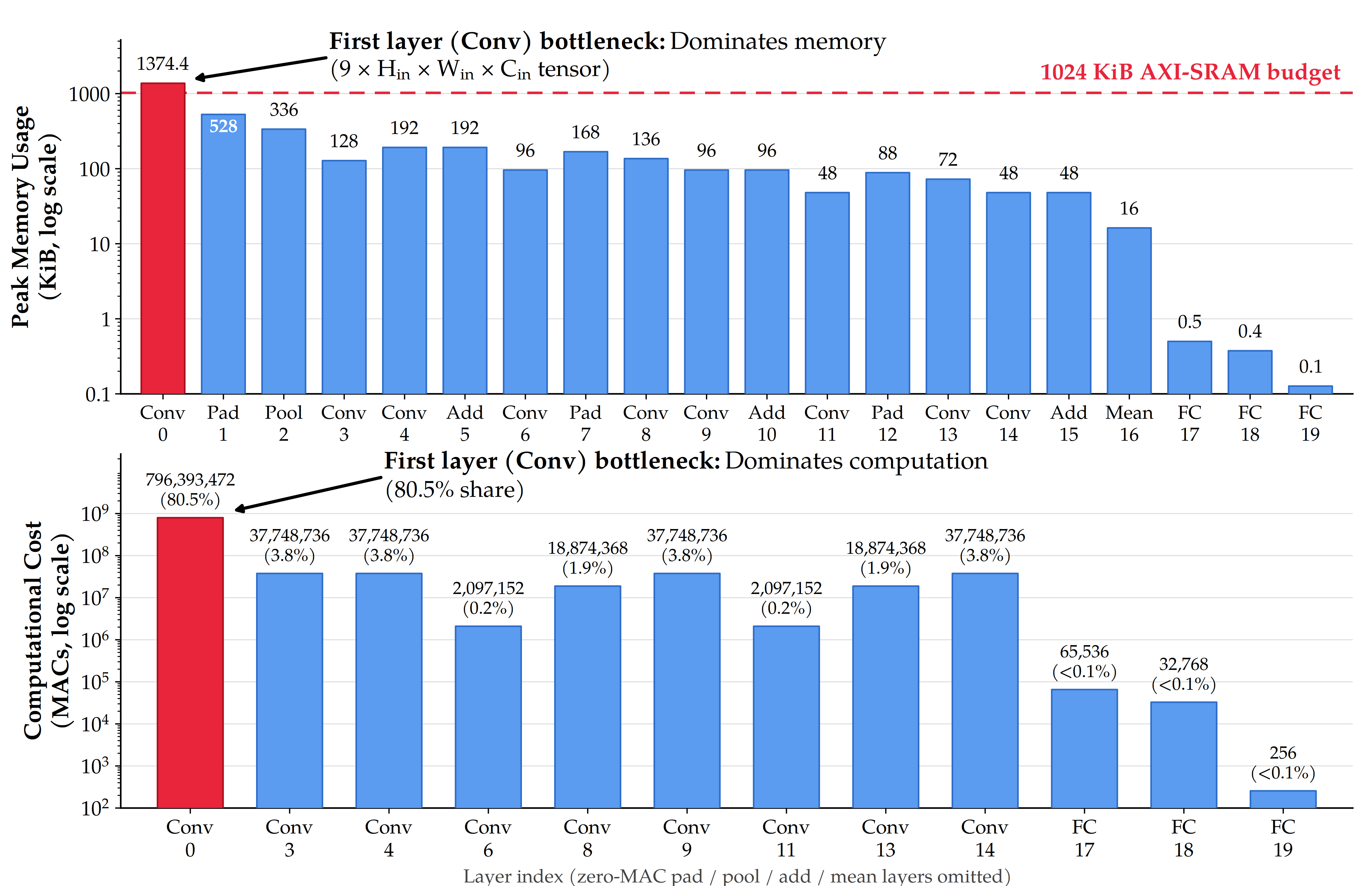}
    \vspace{-7mm}
\caption{\footnotesize Per-layer peak memory (top) and MACs (bottom, zero-MAC layers omitted) for SE-ResNet-8 with 62 electrodes, log scale. The first layer dominates both: 1374.4\,KiB against the 1024\,KiB AXI-SRAM budget, and 80.5\% of all MACs, about 21 times the next-largest layer.}
    \label{fig:motivation-1}
    \vspace{-6mm}
\end{figure}

\vspace{-1mm}
\section{Related Work}
\label{sec:related work}
\vspace{-1mm}

\noindent\textbf{EEG Neural Networks}: EEG models span task-specific CNNs and attention-based architectures~\cite{lawhern2018eegnet,schirrmeister2017deep,song2023eegconformer,altaheri2023atcnet} and pretrained foundation and state-space models~\cite{kostas2021bendr,yang2023biot,jiang2024labram,cui2024neurogpt,wang2025cbramod,ouahidi2025reve,tegon2026femba,wang2025eegmamba}. Despite their diverse backbones, front-end cost often scales with electrode count $C$. In CNNs, early temporal or time--frequency activations grow approximately linearly with $C$, creating peak-memory and compute bottlenecks; in patch-based models, $N_t$ patches per electrode produce $\mathcal{O}(C N_t)$ tokens~\cite{yang2023biot,jiang2024labram}. Unlike approaches that primarily optimize the model architecture, EEG2MCU co-designs the sensing montage and channel-dependent front end through saliency-guided electrode selection, streamed input generation, and memory-constrained first-layer execution, providing a common optimization across CNN, Transformer, and state-space models.

\noindent\textbf{Deployment Frameworks}: MCU deployment is enabled by optimized libraries and runtimes such as CMSIS-NN, TensorFlow Lite Micro, and PULP-NN~\cite{lai2018cmsisnn,david2021tflm,garofalo2020pulpnn}. Memory-aware frameworks further improve feasibility: MCUNet co-designs architectures and execution schedules~\cite{lin2020mcunet}, MCUNetV2 uses patch-based execution to reduce early-layer activations~\cite{lin2021mcunetv2}, DORY automates tiling and transfers across hierarchical memories~\cite{burrello2021dory}, and TinyOps partitions data between internal and external memory using DMA~\cite{sadiq2022tinyops}. MLonMCU supports retargeting and benchmarking across TinyML stacks~\cite{vankempen2023mlonmcu}. Complementing these execution-focused approaches, EEG2MCU co-designs electrode selection, streamed representation generation, and tiled first-layer execution to reduce channel-dependent work and avoid materializing the expanded EEG input across different downstream model families.

\section{Proposed Method}
\label{sec:proposed method}

EEG2MCU optimizes the channel-dependent EEG front end across three coupled
levels. The \emph{sensing level} sets how many electrode streams are
acquired; the \emph{representation level} retains the compact STFT and
expands it only as the executing operator requires; the \emph{execution
level} partitions the first convolution to available SRAM and schedules it
for locality across the MCU memory hierarchy. These levels cannot be
optimized independently: reducing electrode count changes the first-layer
working set, tiling creates SRAM headroom for faster tensor placement, and
streamed generation removes an allocation that would otherwise dominate the
memory budget.

\begin{figure}
    \centering
    \includegraphics[width=0.8\linewidth]{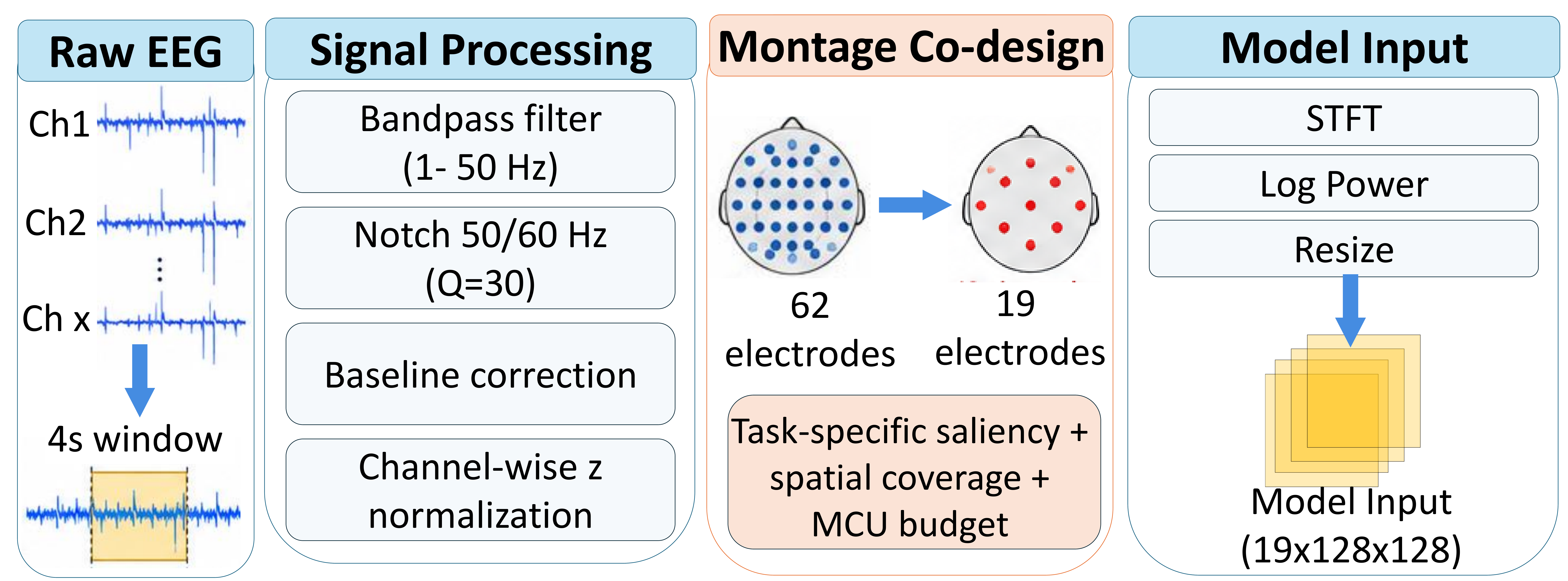}
    \vspace{-4mm}
    \caption{\footnotesize EEG2MCU preprocessing pipeline. Raw multichannel EEG is
    segmented into 4-s windows, filtered and normalized per channel,
    and aligned to the 19-electrode deployment montage to form the
    $19\times128\times128$ model input. }
    \label{fig:sigprocess}
    \vspace{-6mm}
\end{figure}

\vspace{-1mm}
\subsection{Signal Representation and Deployment Backbones}
\label{sec:preprocessing_model}
\vspace{-1mm}

We build on the EEG preprocessing pipeline in~\cite{chowdhury2026bridge}
and redesign its sensing front end for resource-constrained deployment. As
shown in Fig.~\ref{fig:sigprocess}, each recording is segmented into
non-overlapping 4-s windows, and each channel is bandpass filtered to
1--50~Hz, notch filtered at 50 or 60~Hz ($Q=30$), baseline corrected, and
z-normalized. An STFT~\cite{allen1977stft} with a 256-sample Hann window
and 75\% overlap gives a power spectrogram $S_c$ for each electrode $c$,
which is kept to 1--50~Hz and log-scaled as
$\widehat{S}_c=\log_{10}(S_c+10^{-10})$. We resize the spectrogram to
$128\times128$ with endpoint-aligned bilinear interpolation, which keeps
the outermost bins on the grid edges. 

\begin{figure}
    \centering
    \includegraphics[width=0.8\linewidth]{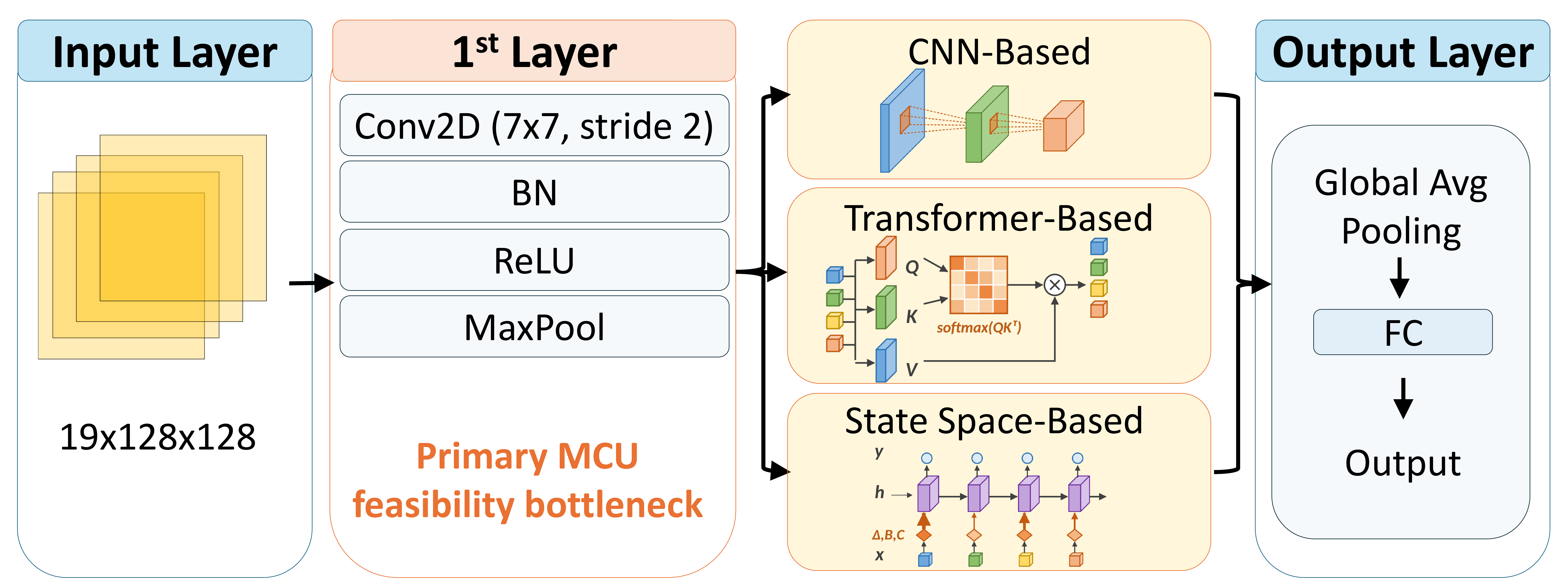}
    \vspace{-4mm}
    \caption{Evaluated model families. All models share a common first
    layer that consumes the full multi-electrode input
    and constitutes the primary MCU feasibility bottleneck, followed by
    a CNN/Transformer/Mamba backbone and a classification head.}
    \label{fig:models}
    \vspace{-4mm}
\end{figure}

The electrode-dependent cost is concentrated in the first layer, the only
layer that reads all $C$ channels (Fig.~\ref{fig:motivation-1}). For deployment, we therefore place a common first layer in front of CNN,
Transformer, and Mamba backbones (Fig.~\ref{fig:models}), so the same
montage-aware and MCU optimizations apply to all of them. Once this layer
fits the MCU memory and latency budget, the remaining layers run with
their standard operators. If the full model still exceeds the node budget, the remaining layers run
on a nearby resource-rich hub. The wearable runs the first
layer and the early backbone layers, where the activations are already
compressed. For EfficientFormer this is a 2.75~KiB tensor at the stage-4 FFN boundary, about
$110\times$ smaller than the model input; for Mamba it is a 5.0~KiB tensor after
the third bidirectional selective-scan block, about $61\times$ smaller. 
Low-power communication modalities such as Wi-R can make this transfer inexpensive:
reported energy is below 100~pJ/bit, compared with more than 15~nJ/bit for
Bluetooth~\cite{WiRTechn50:online}. 
The hub has
much more memory than the MCU, so the remaining layers fit there and the
hub completes the inference. 

\vspace{-1mm}
\subsection{Montage-Aware Streaming Input Generation}
\label{sec:streaming_input}
\vspace{-1mm}

\noindent\textbf{Saliency-guided montage co-design.}
EEG2MCU therefore treats the electrode montage itself as an
algorithm--hardware co-design variable. We first determine a
hardware-feasible electrode budget according to the memory and latency
constraints of the target MCU. Within this budget, electrode-level
sensitivity analysis of the full-channel model is combined with spatial
coverage constraints to retain sensing locations that are both
task-relevant and spatially representative. The deployment montage is selected by maximizing
electrode-level relevance subject to spatial coverage:
\begin{equation}
\mathcal{S}^{*}
=\arg\max_{\mathcal{S}}
\sum_{c\in\mathcal{S}} r_c \ \
\mathrm{s.t.}\ \
|\mathcal{S}|=C_{\mathrm{budget}},
\ \ \mathcal{S}\in\mathcal{C}_{\mathrm{cov}},
\label{eq:montage}
\end{equation}
where $r_c$ denotes electrode saliency computed from
the trained full-channel model using only the training
split, and $\mathcal{C}_{\mathrm{cov}}$ denotes the set
of montages satisfying the spatial-coverage constraint.
We approximate this selection greedily, prioritizing
salient electrodes while maintaining scalp coverage.
We set $C_{\mathrm{budget}}=19$ to enable comparison
with the standard 19-electrode montage while meeting
the 4-s inference window, which all evaluated
62-electrode configurations exceed
(Table~\ref{tab:mcu_breakdown}).
The resulting montage reduces the channel-dependent
first-layer workload by $3.26\times$.

\begin{figure}
    \centering \includegraphics[width=1\linewidth]{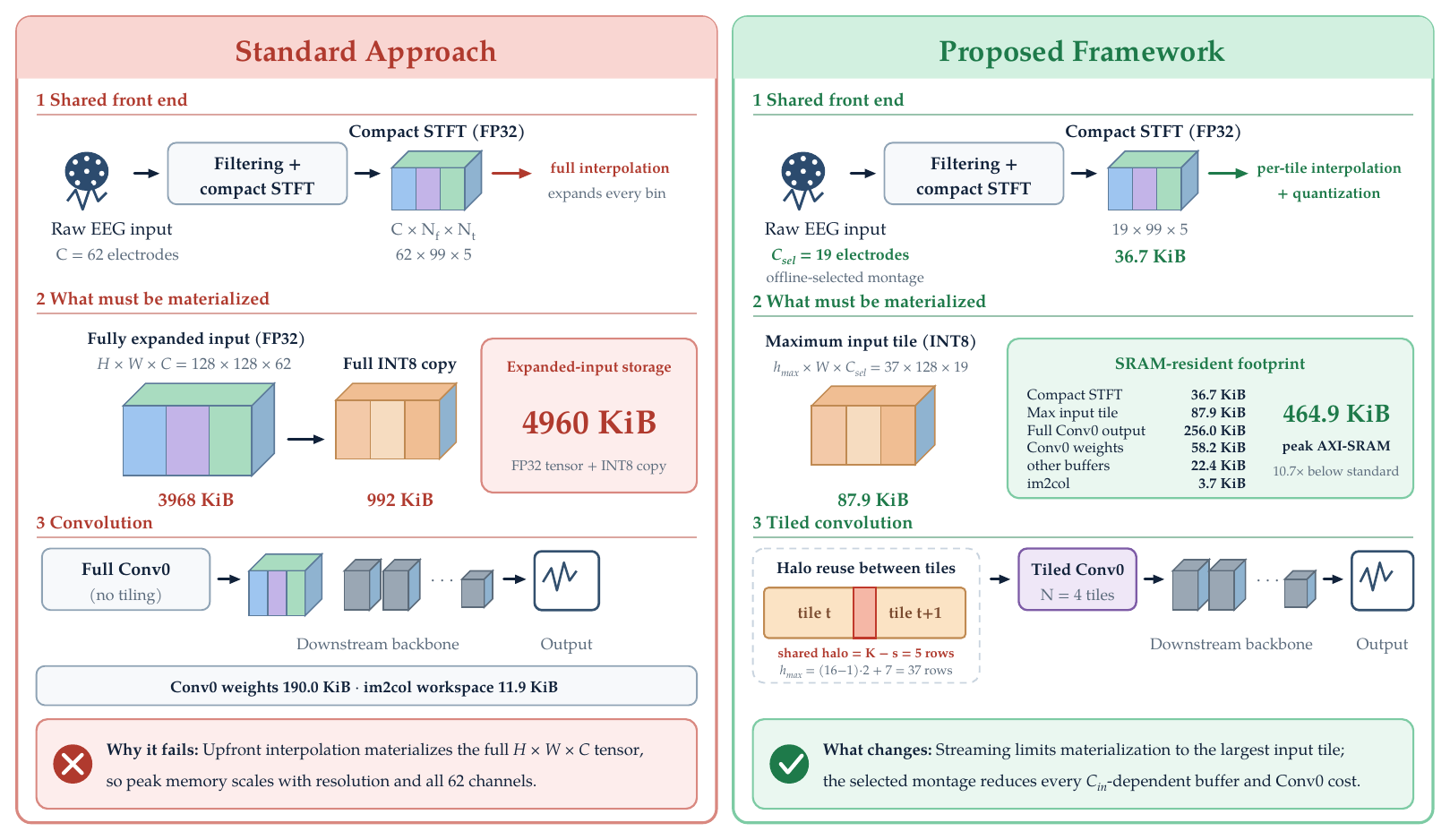}
    \vspace{-8mm}
\caption{\footnotesize Standard (left) versus streamed tiled (right) first-layer execution. Per-tile interpolation and quantization replace full FP32/INT8 input materialization with a bounded INT8 tile buffer.}
    \label{fig:sigprocess-i}
    \vspace{-4mm}
\end{figure}

\noindent\textbf{On-demand representation generation.}
Montage reduction decreases front-end computation, but conventional
preprocessing still constructs the complete high-resolution
time--frequency representation before inference. This creates a large
temporary activation whose lifetime extends until the first layer begins
execution. As illustrated in Fig.~\ref{fig:sigprocess-i}, EEG2MCU instead retains the compact post-STFT representation and
generates only the region required by the currently executing
first-layer tile:
\begin{equation}
\underbrace{C_{\mathrm{sel}}\times N_f\times N_t}_{\text{compact FP32}}
\;\longrightarrow\;
\underbrace{h\times W\times C_{\mathrm{sel}}}_{\text{INT8 tile}},
\label{eq:stream}
\end{equation}
where $C_{\mathrm{sel}}$ is the selected montage size, $N_f$ and $N_t$
are the retained frequency bins and time frames, $H$ and $W$ are the model input dimensions, and $h$ is the number of input rows required by the current
convolution tile. Spatial interpolation and quantization are fused with
tile generation, and the resulting values are written directly into the
first-layer input buffer. The full expanded representation therefore
exists only \emph{logically} and is never simultaneously materialized in
SRAM. Note that this formulation is independent of a specific input
resolution or electrode count.
Conventional materialization requires $M_{\mathrm{full}}=
H W C_{\mathrm{sel}} b_a$, whereas streamed execution requires only $M_{\mathrm{stream}}=
h W C_{\mathrm{sel}} b_a+C_{\mathrm{sel}}N_fN_t b_s,$ where $h\ll H$ is the currently required tile height, and $b_a$ and $b_s$ are the activation and storage precisions respectively. EEG2MCU therefore exchanges a long-lived expanded activation for a compact persistent representation and a short-lived tile
buffer.

\noindent\textbf{Efficient interpolation and fused quantization.}
Let $\widehat{S}_{i,j}$
denote the compact FP32 STFT value at frequency bin $i$ and time bin $j$ for a
given electrode. A location $(r,u)$ in the expanded $H\times W$ model input maps to $y=r\frac{N_f-1}{H-1},$ and $x=u\frac{N_t-1}{W-1}$. Because the source and target dimensions are fixed at deployment, the indices $i=\lfloor y\rfloor$ and $j=\lfloor x\rfloor$ and $f_y=y-i$ and $f_x=x-j$ are precomputed during initialization; upper-neighbor indices are clamped at the boundaries. For each electrode, bilinear interpolation is implemented as two successive one-dimensional interpolations:
\begin{equation}
a_j=(1-f_y)\widehat{S}_{i,j}+f_y\widehat{S}_{i+1,j}, \ \
v_j=(1-f_x)a_j+f_xa_{j+1}.
\label{eq:separable}
\end{equation}
where $a_{j}$ is the intermediate frequency-axis interpolation and $v_{j}$ is the final interpolated FP32 value. For each output row and electrode, intermediate values are reused as the output columns advance, reducing arithmetic and source-memory accesses.

The generator follows a row--electrode--column traversal and writes each result directly into a reusable INT8 buffer for the current tile, stored in NHWC (batch--height--width--channel) order as required by the first-layer convolution. Each interpolated value is immediately quantized as
\begin{equation}
q_{j}=
\mathrm{clip}\left(
\mathrm{round}\left(v_{j}s_{\mathrm{in}}^{-1}\right)
+z_{\mathrm{in}},-128,127
\right),
\end{equation}
where $s_{\mathrm{in}}$ and $z_{\mathrm{in}}$ are the calibrated input scale and zero point. Interpolation remains in floating point until the final rounding, $s_{\mathrm{in}}^{-1}$ is precomputed to avoid per-element division, and inputs known to be zero are written directly as $z_{\mathrm{in}}$. This producer--consumer pipeline generates only the elements required by the current first-layer tile and consumes them immediately.

\subsection{Memory-Constrained INT8 First-Layer Execution}
\label{sec:tiling}

\begin{figure}
    \centering \includegraphics[width=1\linewidth]{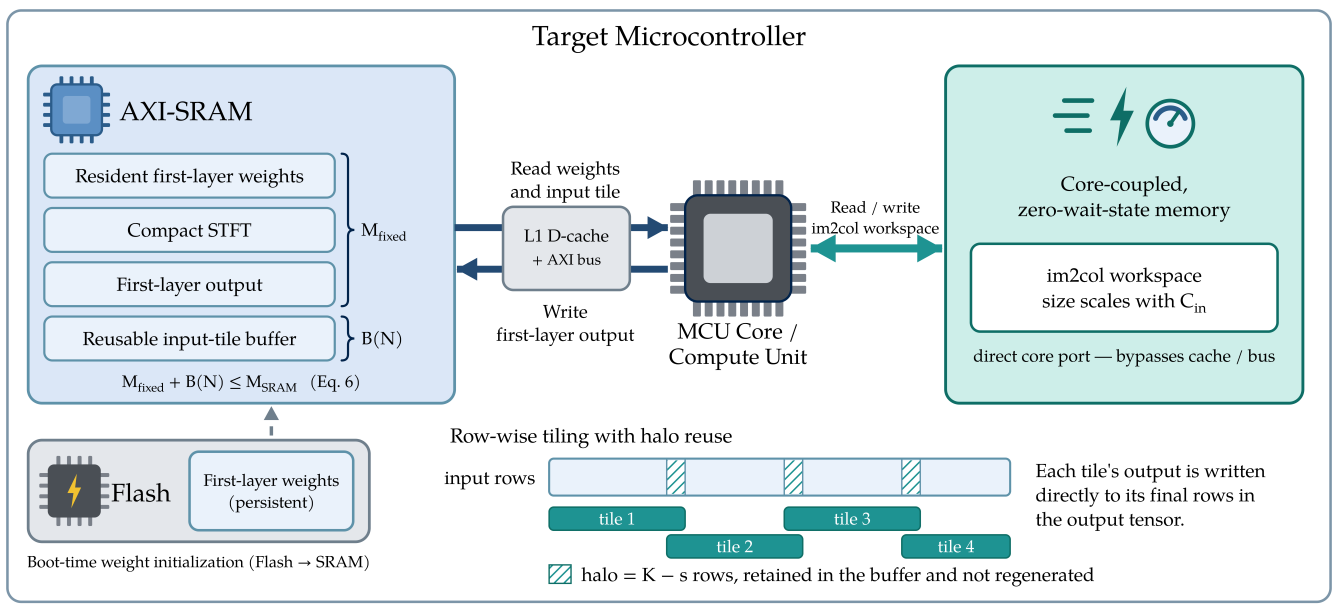}
    \vspace{-9mm}
\caption{\footnotesize EEG2MCU first-layer memory hierarchy and row-wise tiling. Weights are copied once from Flash to AXI-SRAM, the im2col workspace is placed in DTCM, and the minimum feasible tile count is selected under the SRAM budget; halo rows are reused across adjacent tiles.}
    \label{fig:first-layer-tiling}
    \vspace{-5mm}
\end{figure}

After montage reduction and streamed representation generation, the
first layer remains the primary activation-memory bottleneck. EEG2MCU
therefore partitions this layer spatially and executes the resulting
tiles sequentially using optimized CMSIS-NN INT8
kernels~\cite{lai2018cmsisnn}. Related tiled and patch-based execution
schemes partition entire layer stacks for accelerators, distributed
edge devices, or MCUs~\cite{alwani2016fused,zhao2018deepthings,
lin2021mcunetv2,burrello2021dory}; EEG2MCU instead applies tiling only
to the channel-dependent first layer and directly couples it with the
streamed input generator, as illustrated
in Fig.~\ref{fig:first-layer-tiling}.

\noindent\textbf{Row-wise tiling.}
Consider a convolution with kernel height $K$, stride $s$, padding $p$,
input height $H_{\mathrm{in}}$, input width $W_{\mathrm{in}}$, and
$C_{\mathrm{in}}$ input channels. EEG2MCU partitions the
$H_{\mathrm{out}}$ output rows into $N$ contiguous tiles, which leaves
the arithmetic of every output element unchanged. A tile covering
output rows $[r_{\mathrm{s}}, r_{\mathrm{e}}]$ requires the input rows $\left[\, s\,r_{\mathrm{s}}-p,\;\; s\,r_{\mathrm{e}}-p+K-1 \,\right]
\cap
\left[\, 0,\; H_{\mathrm{in}}-1 \,\right]$. Because CMSIS-NN applies padding internally rather than requiring a
pre-padded input, the clipped rows are supplied as
tile-local top and bottom padding. Consequently, only the first and last
tiles receive vertical padding, while all interior boundaries fall
strictly within the input tensor. Adjacent tiles overlap by $K-s$ input
rows, and the maximum input-tile buffer is bounded by
\begin{equation}
B(N)=
\left(
\left\lceil
\frac{H_{\mathrm{out}}}{N}
\right\rceil s
+K-s
\right)
W_{\mathrm{in}}C_{\mathrm{in}}b_a ,
\label{eq:buffer}
\end{equation}
where $b_a$ denotes the number of bytes per activation. Tiles are
processed sequentially using a single reusable input buffer, while each
output region is written directly to its final location. Increasing
$N$ therefore reduces the peak input footprint without changing the
underlying convolution.

\noindent\textbf{Quantization and numerical consistency.}
We use symmetric per-output-channel INT8 weight
quantization and calibrated INT8 activations, with
CMSIS-NN requantization parameters derived from
$m_c=s_{\mathrm{in}}s_{w,c}/s_{\mathrm{out}}$.
During model conversion, we reconcile signed
requantization shifts and convolution padding
to prevent numerical mismatches between the
reference model and tiled MCU execution.

\noindent\textbf{Feasibility-driven tile selection.}
Finer tiling is not always beneficial because it increases kernel
invocations and exposes additional tile-boundary overhead. EEG2MCU
therefore selects the minimum number of tiles required to satisfy the
runtime-memory constraint,
\begin{equation}
N_{\min}
=
\min
\left\{
N:
M_{\mathrm{fixed}}+B(N)
\leq M_{\mathrm{SRAM}}
\right\},
\label{eq:nstar}
\end{equation}
where $M_{\mathrm{fixed}}$ includes the output
activation, convolution workspace, SRAM-resident
weights, and other simultaneously resident buffers.
The minimum feasible tile count establishes the
capacity boundary; additional tiling may be selected
to provide SRAM headroom when its measured latency
and energy overhead are acceptable. Filtering and
STFT are computed once per EEG window, so tiling
changes only the first-layer execution schedule
and working-set size without modifying the
convolution arithmetic.

\begin{table*}[t]
\centering
\caption{MCU Execution Breakdown of Streamed Generation and Hierarchy-Aware Execution}
\label{tab:mcu_breakdown}
\vspace{-8pt}
{\scriptsize ``---'' denotes configurations whose working set exceeds the SRAM budget and cannot be deployed.\par}
\scriptsize
\setlength{\tabcolsep}{0pt}
\renewcommand{\arraystretch}{1.05}
\begin{tabular*}{\textwidth}{@{\extracolsep{\fill}} c c c c c r r r r r r r r r r r @{}}
\toprule
\multirow{2}{*}{\textbf{Ch}} & \multirow{2}{*}{\textbf{Tiles}} & \multicolumn{3}{c}{\textbf{{**}Hierarchy-Aware Execution}} & \textbf{Streamed} & \textbf{Layer 0} & \multicolumn{4}{c}{\textbf{Latency} (ms)} & \multicolumn{4}{c}{\textbf{Energy} (mJ)} & \textbf{Peak} \\
\cmidrule{3-5} \cmidrule{8-11} \cmidrule{12-15}
 & & \textbf{Weight Res.} & \textbf{DTCM} & \textbf{Halo} & \textbf{Gen.} (ms) & (ms) & \textbf{RN 8} & \textbf{SE} & \textbf{Eff.\textsuperscript{*}} & \textbf{Mamba\textsuperscript{*}} & \textbf{RN 8} & \textbf{SE} & \textbf{Eff.\textsuperscript{*}} & \textbf{Mamba\textsuperscript{*}} & (KiB) \\
\midrule
62 & 1  & \xmark & \xmark & \xmark & ---   & ---    & ---     & ---     & ---     & ---     & ---   & ---   & ---   & ---   & 1{,}374.4 \\
62 & 2  & \xmark & \xmark & \xmark & 30.2  & 4913.6 & 5492.7  & 5554.8  & 5492.0  & 5629.4  & 791.9 & 801.6 & 788.5 & 806.1 & 850.0 \\
62 & 2  & \cmark & \xmark & \xmark & ---   & ---    & ---     & ---     & ---     & ---     & ---   & ---   & ---   & ---   & 1{,}077.4 \\
\midrule
62 & 4  & \xmark & \xmark & \xmark & 30.2  & 4924.3 & 5503.4  & 5565.5  & 5502.7  & 5640.1  & 795.3 & 804.9 & 791.8 & 809.5 & 617.5 \\
62 & 4  & \cmark & \xmark & \xmark & 30.2  & 4622.3 & 5201.4  & 5263.5  & 5200.7  & 5338.1  & 700.2 & 709.9 & 696.8 & 714.5 & 826.7 \\
62 & 4  & \cmark & \cmark & \xmark & 30.2  & 4398.4 & 4977.5  & 5039.6  & 4976.8  & 5114.2  & 699.6 & 709.2 & 696.2 & 713.8 & 795.5 \\
\midrule
\textbf{62} & \textbf{4} & \textbf{\cmark} & \textbf{\cmark} & \textbf{\cmark} & \textbf{29.3} & \textbf{4344.4} & \textbf{4923.5} & \textbf{4985.6} & \textbf{4922.8} & \textbf{5060.2} & \textbf{692.0} & \textbf{701.6} & \textbf{688.6} & \textbf{706.2} & \textbf{795.5} \\
62 & 8  & \cmark & \cmark & \cmark & 29.3  & 4346.1 & 4925.1  & 4987.2  & 4924.4  & 5061.8  & 692.6 & 702.2 & 689.1 & 706.8 & 671.5 \\
62 & 16 & \cmark & \cmark & \cmark & 29.3  & 4349.4 & 4928.4  & 4990.5  & 4927.7  & 5065.1  & 693.2 & 702.9 & 689.8 & 707.4 & 609.5 \\
\midrule
19 & 1  & \xmark & \xmark & \xmark & 18.0  & 1419.5 & 1998.5  & 2060.6  & 1997.8  & 2135.3  & 296.7 & 306.4 & 293.3 & 310.9 & 601.3 \\
19 & 1  & \cmark & \xmark & \xmark & 18.0  & 1369.2 & 1948.3  & 2010.4  & 1947.6  & 2085.0  & 293.3 & 302.9 & 289.9 & 307.5 & 659.9 \\
19 & 1  & \cmark & \cmark & \xmark & 17.9  & 1351.0 & 1930.1  & 1992.2  & 1929.4  & 2066.8  & 278.6 & 288.3 & 275.2 & 292.9 & 665.6 \\
\textbf{19} & \textbf{4} & \textbf{\cmark} & \textbf{\cmark} & \textbf{\cmark} & \textbf{19.4} & \textbf{1339.0} & \textbf{1918.1} & \textbf{1980.2} & \textbf{1917.4} & \textbf{2054.8} & \textbf{279.4} & \textbf{289.1} & \textbf{276.0} & \textbf{293.7} & \textbf{464.9} \\
\bottomrule
\end{tabular*}
\vspace{2pt}
\begin{minipage}{\textwidth}
\scriptsize\textsuperscript{*}EfficientFormer and Mamba exceed the 2\,MB Flash budget at INT8 and are deployed on-device only up to the stage-4 FFN boundary and the third of six bidirectional selective-scan blocks, respectively; the remaining layers execute on the hub. RN8 and SE are deployed in full. \par
{\scriptsize {**}Hierarchy-aware execution ablates: Weight Res.\ (SRAM weight residency), DTCM (DTCM workspace routing), and Halo (halo buffer reuse).\par}
\end{minipage}
\vspace{-7mm}
\end{table*}

\begin{table}[!t]
\centering
\caption{Model Performance on SEED Across Montage Configurations}
\label{tab:model_comparison}
\vspace{-2mm}
\scriptsize
\setlength{\tabcolsep}{2.3pt}
\renewcommand{\arraystretch}{1.08}

\resizebox{\columnwidth}{!}{%
\begin{tabular}{@{}l *{9}{c}@{}}
\toprule
& \multicolumn{3}{c}{\textbf{SE-ResNet8}}
& \multicolumn{3}{c}{\textbf{EfficientFormer~\cite{li2022efficientformer}}}
& \multicolumn{3}{c}{\textbf{Mamba~\cite{gui2024eegmamba}}} \\
\cmidrule(lr){2-4}
\cmidrule(lr){5-7}
\cmidrule(lr){8-10}

\textbf{No. Electrodes}
& \textbf{Acc.} & \textbf{F1} & \textbf{Params}
& \textbf{Acc.} & \textbf{F1} & \textbf{Params}
& \textbf{Acc.} & \textbf{F1} & \textbf{Params} \\
\midrule

62
& 0.781 & 0.782 & 1.5M
& 0.769 & 0.767 & 3M
&  0.785 & 0.772  & 2.5M \\        

19 (Std 10--20)
& 0.619 & 0.618 & 1.5M
& 0.625 & 0.614 & 3M
& 0.685 & 0.668 & 2.5M \\

19 (Saliency-guided)
& 0.734 & 0.730 & 1.5M
& 0.736 & 0.736 & 3M
& 0.737 & 0.735 & 2.5M \\
\bottomrule
\end{tabular}%
}
\vspace{-7mm}
\end{table}

\vspace{-1mm}
\subsection{Locality- and Memory-Hierarchy-Aware Execution}
\label{sec:locality}
\vspace{-1mm}

Satisfying the SRAM capacity constraint does not by itself
guarantee efficient execution. In the evaluated CMSIS-NN
implementation, the first convolution repeatedly accesses
its weights and \texttt{im2col} workspace, making latency
and energy sensitive to where these tensors reside in the
MCU memory hierarchy~\cite{lai2018cmsisnn}.
EEG2MCU therefore combines halo-preserving traversal
with access-aware placement across Flash, SRAM, and
tightly coupled memory to reduce redundant computation
and data movement.

\noindent\textbf{Halo-preserving spatial traversal.}
Adjacent convolution tiles share $(K-s)$ input rows because
their receptive fields overlap. A naive tiled implementation
regenerates these shared regions independently, increasing
interpolation and quantization overhead with the number
of tiles. EEG2MCU instead processes tiles in spatial order
and retains overlapping rows as a halo in the reusable
input buffer. When advancing to the next tile, only newly
required rows are generated, interpolated, and quantized.

This reduces the total number of generated rows to
$H_{\mathrm{in}}$, independent of the tile count. Because
the retained halo forms part of the next tile's input,
it resides within $B(N)$ and does not increase peak
buffer memory. The resulting execution preserves
numerical equivalence with full-tensor inference.

\noindent\textbf{Hierarchy-aware tensor placement.}
\label{sec:memory_optimization}
EEG2MCU places tensors by capacity and access intensity
rather than default linker placement. First-layer weights
are copied once from Flash to SRAM at initialization,
avoiding repeated Flash accesses while retaining the
Flash image as persistent model storage. The SRAM copy
increases $M_{\mathrm{fixed}}$ in Eq.~\eqref{eq:nstar},
enabled by the headroom from spatial tiling.
The frequently accessed CMSIS-NN \texttt{im2col}
workspace resides in zero-wait-state DTCM, larger
activations in SRAM, and infrequently accessed
constants in Flash. Buffers are cache-line aligned,
with SRAM configured as cacheable~\cite{stAN4839},
reducing repeated-access costs. 
Reported peak memory is the total on-chip footprint; the \texttt{im2col} workspace
resides in the 128\,KiB DTCM rather than AXI-SRAM, and no off-chip memory is
required at any point.

\noindent\textbf{Joint schedule selection.}
Spatial tiling and memory placement are coupled:
finer tiling reduces the activation working set,
creating SRAM headroom for faster weight placement,
while halo reuse limits the resulting regeneration
overhead. EEG2MCU therefore jointly selects the
number of tiles $N$ and execution configuration
$\mathbf{p}=(p_w,p_d,p_h)$, where the binary variables
indicate SRAM weight residency, DTCM workspace
placement, and halo reuse, respectively. Because model dimensions and hardware resources are
fixed after deployment, candidate configurations are
evaluated offline on the target MCU. The selected
configuration minimizes measured end-to-end latency
subject to the available memory budget:
\begin{equation}
(N^*,\mathbf{p}^*){=}\arg\min_{N,\mathbf{p}}
T_{\mathrm{e2e}}(N,\mathbf{p}) 
\ \ \mathrm{s.t.} \ \
M_{\mathrm{peak}}(N,\mathbf{p})
{\leq}M_{\mathrm{SRAM}}, \notag
\label{eq:schedule}
\end{equation}
where $M_{\mathrm{peak}}$ accounts for the
simultaneously resident input tile, output
activations, weights, workspace, and persistent
runtime state under the selected memory placement.
This formulation captures the capacity--latency
trade-off between spatial decomposition and
memory-hierarchy utilization. The selected
configuration is fixed at deployment and introduces
no runtime search overhead.

\vspace{-1mm}
\section{Experiments}
\label{sec:experiments}
\vspace{-1mm}

\subsection{Experimental Setup} 
\label{sec:experimental_setup}
\vspace{-1mm}

\noindent\textbf{Target platform and measurement}: We evaluate our framework on an STM32H7A3ZI (Cortex-M7 at 280\,MHz, 1\,MB AXI-SRAM, and 2\,MB Flash). Inference latency is measured with the hardware DWT cycle counter, and energy is measured with an STLINK-V3PWR kit. Each configuration is executed 5 times; the first iteration is discarded to remove cache warm-up effects, and we report the mean of the remaining iterations.

\noindent\textbf{Benchmark models}: We evaluate ResNet-8 (RN8), SE-ResNet-8 (SE), EfficientFormer-V2 (Eff), and Mamba, spanning convolutional, Transformer, and state-space backbones. All share the first layer in Fig.~\ref{fig:models}: a $7\times7$, stride-2 convolution with 64 output channels, followed by batch normalization, ReLU, and $3\times3$ max pooling. SE comprises three residual stages~\cite{he2016resnet} with squeeze-and-excitation blocks~\cite{hu2018senet}, global average pooling, and a classifier; RN8 omits the squeeze-and-excitation blocks and is used only for hardware characterization. Eff adopts EfficientFormer-V2-S0 stages~\cite{li2022efficientformer}, with attention-bias tables matched to the first-layer output resolution. Mamba follows bidirectional EEGMamba~\cite{gui2024eegmamba}, projecting each temporal position into a token processed by six bidirectional selective-scan blocks (width 160, state size 16), followed by temporal mean pooling and a linear classifier. SE is distilled from a self-supervised SE-ResNet-18 teacher~\cite{chowdhury2026bridge} ($\tau{=}4$, $\lambda_{\text{KD}}{=}0.5$) and fine-tuned; Eff and Mamba are trained from scratch. Training is fixed across montages. Accuracy is evaluated on three-class SEED~\cite{zheng2015seed} using a random 80/20 split of 4-s windows. All models are quantized to INT8. 


 \vspace{-1mm}   
\subsection{Performance Analysis}
 \vspace{-1mm}

\noindent\textbf{Effects of montage selection on accuracy}: 
Table~\ref{tab:model_comparison} compares 62 electrodes with standard 10--20 and saliency-guided 19-electrode montages. The standard montage loses 13--21\% accuracy, whereas ours retains 94--96\% of full-montage accuracy and reduces first-layer cost by $3.26\times$. Unlike the general-purpose 10--20 layout, SEED benefits from lateral-temporal and prefrontal beta/gamma activity; the standard layout retains only four of the 12 most informative lateral-temporal electrodes~\cite{zheng2015seed}. In contrast, our method balances training-only saliency with spatial coverage (see Eq.~\eqref{eq:montage}). Lower resolutions lose spectral detail: with SE and 19 electrodes, $64\times64$ and $32\times32$ achieve 69.72\% and 66.82\% accuracy. Since streamed tiling fits $128\times128$ within SRAM, we retain this resolution.


\begin{table}[!t]
\centering
\caption{Comparison with MCU deployment frameworks and prior EEG-on-MCU deployment}
\label{tab:framework_comparison}
\vspace{-18pt}
\vspace{2pt}
\begin{minipage}{\columnwidth}
\end{minipage}\par
\vspace{2pt}
\resizebox{\columnwidth}{!}{%
\setlength{\tabcolsep}{3.5pt}
\renewcommand{\arraystretch}{1.18}
\begin{tabular}{@{} l l c c c c c @{}}
\toprule
\textbf{Framework} & \textbf{Target}
  & \textbf{\shortstack{No ext.\\mem.\,\textsuperscript{}}}
  & \textbf{\shortstack{On-demand\\input gen.\,\textsuperscript{}}}
  & \textbf{\shortstack{Sensing\\in scope\,\textsuperscript{[a]}}}
  & \textbf{\shortstack{Arb. kern.\\geometry\,\textsuperscript{[b]}}}
  & \textbf{\shortstack{Cache/TCM\\placement\,\textsuperscript{[c]}}} \\
\midrule
DORY~\cite{burrello2021dory}         & GAP-8 + ext.\ L3  & \xmark & \xmark & \xmark & \cmark & \xmark \\
TinyOps~\cite{sadiq2022tinyops}      & M4/M7 + SDRAM     & \xmark & \xmark & \xmark & \cmark & \xmark \\
TinyEngine~\cite{lin2020mcunet}      & Cortex-M          & \cmark & \xmark & \xmark & \xmark & \xmark \\
MLonMCU~\cite{vankempen2023mlonmcu}  & Retargetable      & \cmark & \xmark & \xmark & \cmark & \xmark \\
\midrule
\textbf{EEG2MCU (ours)}              & \textbf{Cortex-M7} & \textbf{\cmark} & \textbf{\cmark} & \textbf{\cmark} & \textbf{\cmark} & \textbf{\cmark} \\
\bottomrule
\end{tabular}%
}

\vspace{3pt}
\begin{minipage}{\columnwidth}
\scriptsize
\textsuperscript{[a]}\textbf{Sensing in scope}: sensor dimensionality and representation generation are optimization variables.
\textsuperscript{[b]}\textbf{Arb.\ kern.\ geometry}: supports convolution kernel sizes beyond a fixed search space.
\textsuperscript{[c]}\textbf{Cache/TCM placement}: access-aware tensor assignment across a cached Flash/SRAM/DTCM hierarchy.
\end{minipage}


\begin{minipage}{\columnwidth}
\end{minipage}\par
\vspace{2pt}

\resizebox{\columnwidth}{!}{%
\setlength{\tabcolsep}{4.5pt}
\renewcommand{\arraystretch}{1.2}
\begin{tabular}{@{} l l l c c c @{}}
\toprule
\textbf{Work} & \textbf{Network deployed} & \textbf{Target} &
\textbf{\shortstack{Peak SRAM\\(KiB)}} &
\textbf{\shortstack{Latency\\(ms)}} &
\textbf{\shortstack{Energy\\(mJ)}} \\
\midrule
FEMBA\textsuperscript{$\dagger$}~\cite{tegon2026femba}
 & \shortstack[l]{Mamba (22\,ch, 7.8\,M params)}
 & GAP9 + ext.\ L3 mem.
 & --- & 1700 & 75 \\
\midrule
\multirow{1}{*}{\textbf{EEG2MCU (ours)}}
 & {\shortstack[l]{SE-ResNet-8 (1.56\,MB)}}
 & {\shortstack[l]{STM32H7 (Cortex-M7)}}
 & {464.9} & {1980.2} & {289.1} \\
\bottomrule
\end{tabular}%
}
\begin{minipage}{\columnwidth}
\scriptsize
\textsuperscript{*}Each work's published results on its own network, input, and platform;
rows are not directly comparable. ``---'' denotes not reported.
\textsuperscript{$\dagger$}Preprocessing performed offline and excluded from reported latency and energy.
\vspace{-4mm}
\end{minipage}

\end{table}

\begin{figure}
    \centering \includegraphics[width=1\linewidth]{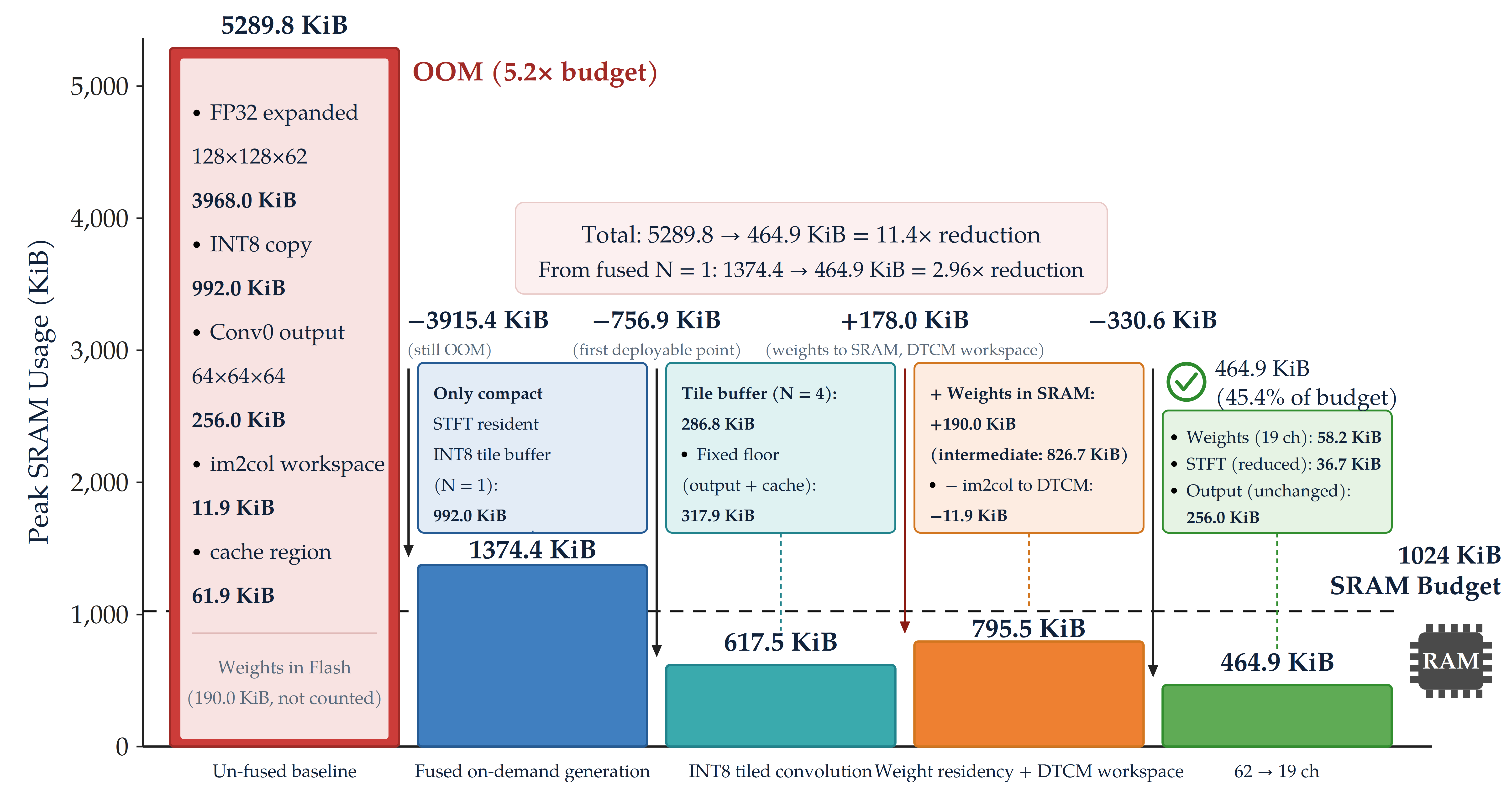}
    \vspace{-8mm}
\caption{\footnotesize Peak SRAM across cumulative front-end optimizations relative to the 1024\,KiB budget, decreasing from 5289.8\,KiB for the unfused 62-channel baseline to 464.9\,KiB (11.38$\times$ reduction) for the optimized 19-channel design. Callouts show the peak-memory breakdown at each stage.}
    \label{fig:mem-waterfall}
    \vspace{-3mm}
\end{figure}

\begin{figure}[!t]
         \centering
         \includegraphics[width=1\linewidth]{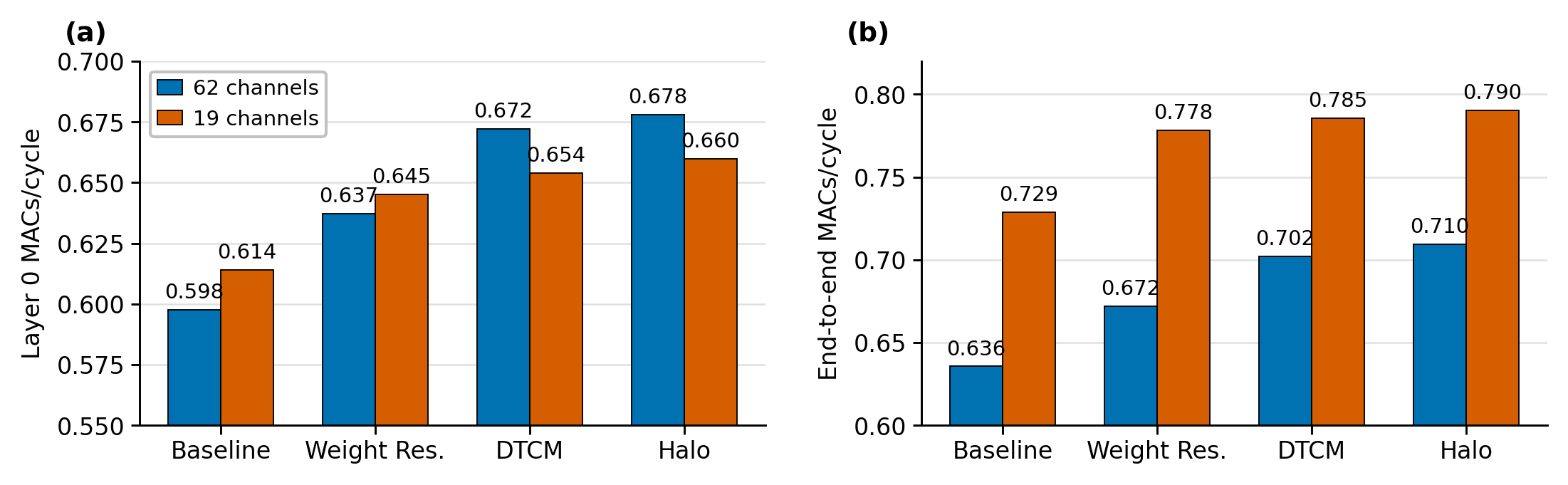}
          \vspace{-8mm}
\caption{\footnotesize MACs/cycle under cumulative SRAM weight residency, DTCM placement, and halo reuse for 62- and 19-channel inputs: (a) layer-0 convolution and (b) end-to-end SE-ResNet-8.}
    \label{fig:MACs/cycle-bar}
    \vspace{-5mm}
    \end{figure}

\noindent\textbf{Comparison with prior deployments.}
Table~\ref{tab:framework_comparison} compares EEG2MCU with existing MCU deployment frameworks. DORY~\cite{burrello2021dory} uses constraint-based layer tiling for GAP-8 DMA-managed scratchpads, streams weights from external L3, and optimizes only beyond the ONNX boundary. TinyEngine~\cite{lin2020mcunet} supports TinyNAS-generated operator geometries, excluding our stem convolution and explicit padding. TinyOps~\cite{sadiq2022tinyops} tiles spatially using external SDRAM, relocating (not eliminating) the working set, while MLonMCU~\cite{vankempen2023mlonmcu} is a benchmarking harness without energy measurements. All four assume a materialized model input, which alone exceeds our MCU's SRAM (Fig.~\ref{fig:mem-waterfall}); downstream tiling cannot recover this memory. Table~\ref{tab:framework_comparison} also compares EEG deployments measured on each work's own network, input, and platform. Most EEG models, including NAS-derived architectures~\cite{10181950,li2023eegnas} and foundation models~\cite{wang2025cbramod,wang2025eegmamba,ouahidi2025reve}, lack MCU deployment. The closest work, FEMBA~\cite{tegon2026femba}, deploys quantized Mamba on GAP9 using scratchpad-specific code generation and tiling, streams weights externally, fixes the montage at 22 channels, and does not support on-demand preprocessing. In contrast, EEG2MCU deploys the backbones on a commodity single-core Arm Cortex-M7 without external memory, fuses interpolation and quantization into first-layer tiling to support preprocessing, and co-optimizes electrode count. 

\noindent\textbf{Effects of streamed generation and row-wise tiling}: The unfused 62-channel front end materializes FP32 and INT8 inputs, requiring 5289.8\,KiB (Fig.~\ref{fig:mem-waterfall}). As shown in Table~\ref{tab:mcu_breakdown}, fusing interpolation and quantization removes the FP32 intermediate and reduces the input footprint by $5.0\times$, although the untiled $N{=}1$ schedule still requires 1374.4\,KiB, exceeding the 1024\,KiB SRAM budget. Streamed tiling reduces the peak to 617.5\,KiB at $N{=}4$ ($55.1\%$), the smallest feasible tile count predicted by Eq.~\eqref{eq:nstar}, since $N{=}2$ requires 1077.4\,KiB with resident weights. It also achieves the lowest latency and energy among feasible schedules. Increasing to $N{=}16$ saves another 186\,KiB for only 5\,ms and 1.3\,mJ. 
At $N=4$, halo reuse eliminates 10.5\% of generated input rows (143 to 128) and reduces streamed-generation latency from 30.2\,ms to 29.3\,ms at 62 electrodes, without additional buffering.

\begin{figure}
    \centering
    \includegraphics[width=1\linewidth]{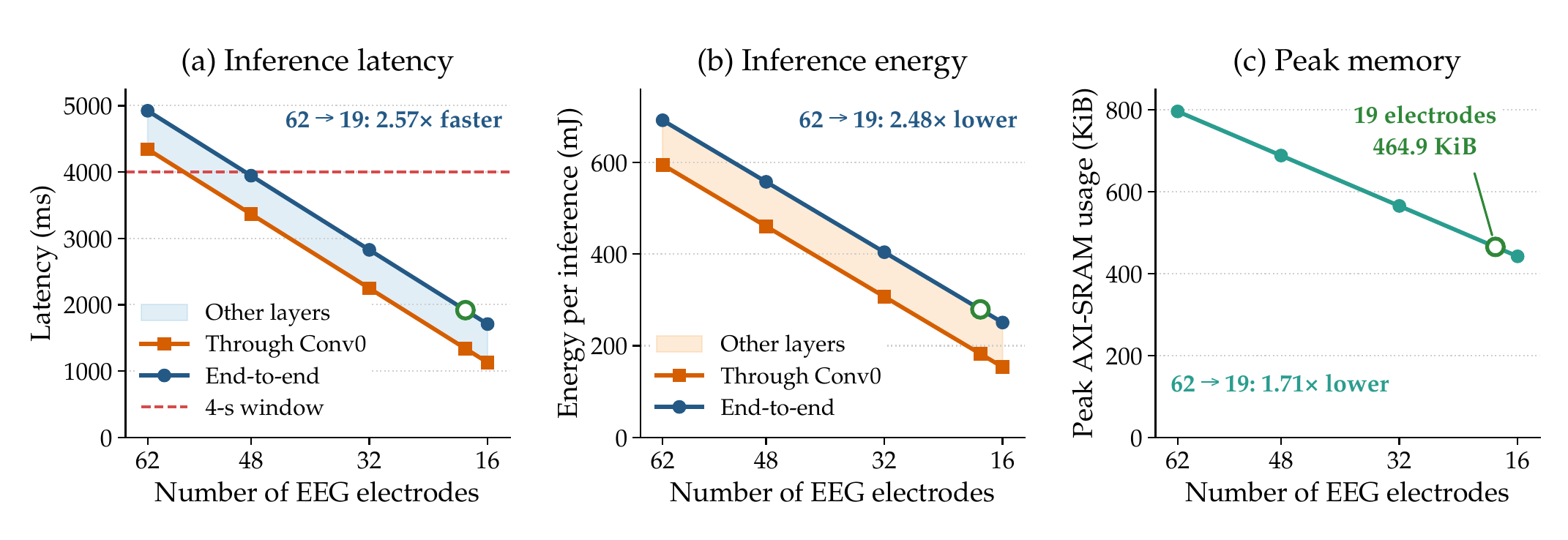}
    \vspace{-8mm}
\caption{\footnotesize Montage-aware ResNet-8 scaling at $N=4$ with all memory optimizations enabled: (a) latency, (b) energy, and (c) peak AXI-SRAM. Versus 62 electrodes, the selected 19-electrode montage reduces these metrics by $2.57\times$, $2.48\times$, and $1.71\times$, respectively (Table~\ref{tab:mcu_breakdown}), while meeting the 4\,s window and 1024\,KiB AXI-SRAM limits.}
    \label{fig:plot-1}
    \vspace{-6mm}
\end{figure}

\noindent\textbf{Effects of memory-hierarchy-aware execution}: Each inference performs $H_{\mathrm{out}}W_{\mathrm{out}}/2=2048$ column evaluations, repeatedly streaming the 190.0\,KiB weight tensor through the 16\,KiB L1 data cache~\cite{stAN4839}, roughly $12\times$ the cache capacity.
SRAM weight residency eliminates Flash stalls, reducing ResNet-8 energy by 12.0\% (795.3\,mJ to 700.2\,mJ), while DTCM workspace placement removes bus contention without changing Flash traffic. 
Combined with halo reuse, this access-aware schedule reduces ResNet-8 latency by 10.5\% and energy by 13.0\% at $N = 4$ for 62 channels.
As shown in Fig.~\ref{fig:MACs/cycle-bar}, all three stages increase layer-0 MACs/cycle from 0.598 to 0.678 ($+13.4\%$) and 0.614 to 0.660 ($+7.5\%$) for 62 and 19 channels, respectively. End-to-end SE-ResNet-8 throughput similarly rises from 0.636 to 0.710 ($+11.6\%$) and 0.729 to 0.790 ($+8.4\%$). Weight residency adds 209.2\,KiB to peak memory, but $N=16$ tiling recovers 186\,KiB, lowering the peak below the unoptimized baseline with negligible latency and energy overhead.

\noindent\textbf{Effects of montage co-design}: Reducing the montage from 62 to 19 electrodes directly lowers activation memory, weight traffic, and MACs (see Fig.~\ref{fig:plot-1}). First-layer weights shrink by $3.26\times$, while signal-processing-to-layer-0 latency falls by $3.24\times$, from 4344.4\,ms to 1339.0\,ms. At $N=4$, peak memory decreases by 41.6\%, from 795.5\,KiB to 464.9\,KiB, the lowest evaluated footprint. Whole-model throughput increases by 11.3\%, from 0.710 to 0.790 MACs/cycle (see Fig.~\ref{fig:MACs/cycle-bar}(b)). 
Consequently, SE-ResNet-8 latency improves by $2.52\times$ (4985.6\,ms to 1980.2\,ms) and energy by $2.43\times$ (701.6\,mJ to 289.1\,mJ). The channel-independent backbone therefore grows from 12.9\% to 32.4\% of end-to-end latency, although the stem remains dominant.
\section{Conclusion}
\vspace{-1mm}

EEG2MCU addresses the channel-dependent EEG front end,
which accounts for 87.1\% of latency and 80.5\% of
computation. Joint electrode selection, streamed INT8
generation, tiling, and memory-aware execution reduce
peak SRAM by $11.38\times$, latency by $2.57\times$,
and energy by $2.48\times$ on a Cortex-M7, retaining
94--96\% of full-montage accuracy. The shared stem
supports CNN, Transformer, and state-space backbones,
but its remaining 67.6\% latency share motivates
further joint stem--backbone optimization.

\clearpage

\begingroup
\linespread{0.91}\selectfont
\bibliographystyle{IEEEtran}
\bibliography{Reference}
\endgroup
\end{document}